\documentclass[sigconf,screen]{acmart}
\AtBeginDocument{%
  }

\usepackage{graphicx}
\usepackage{multirow}
\usepackage{comment}
\usepackage[utf8]{inputenc}
\usepackage{amsthm}\usepackage{float}
\usepackage[table]{xcolor}
\usepackage[edges]{forest}
\usetikzlibrary{arrows.meta}
\renewcommand{\arraystretch}{1.7}
\theoremstyle{definition}
\newtheorem{definition}{Definition}[section]
\usepackage{soul}
\usepackage[normalem]{ulem}

\setcopyright{rightsretained}
\acmDOI{10.1145/3805773.3806001}
\acmYear{2026}
\copyrightyear{2026}
\acmISBN{979-8-4007-2602-6/2026/07}
\acmConference[SecDev '26]{Proceedings of the 2026 ACM Secure Development Conference}{July 5--6, 2026}{Montreal, QC, Canada}
\acmBooktitle{Proceedings of the 2026 ACM Secure Development Conference (SecDev '26), July 5--6, 2026, Montreal, QC, Canada}
\acmSubmissionID{fsesecdev26main-p26-p}
\received{2026-02-05}
\received[accepted]{2026-03-19}

\begin{document}


\title{SoK: A Comprehensive Analysis of the Current Status of Neural Tangent Generalization Attacks with Research Directions}
\titlenote{We acknowledge NSF for partially sponsoring the work under grants~\#2519356, \#2236283, and \#1620868 with its REU. We also thank Cyber Florida for a seed grant and Rapid7 for its endowment to USF.}

\author{Thushari Hapuarachchi}
\orcid{0009-0001-0067-6997}
\affiliation{%
  \institution{University of South Florida}
  \city{Tampa}
  \country{USA}
}
\email{saumya2@usf.edu}

\author{Kaiqi Xiong}
\orcid{0000-0003-2933-8083}
\affiliation{%
  \institution{University of South Florida}
  \city{Tampa}
  \country{USA}
}
\email{xiongk@usf.edu}

\begin{abstract}
There is recently a serious issue that Deep Neural Networks (DNNs) training uses more and more unauthorized data. A clean-label generalization attack, one type of data poisoning attacks, has been suggested to address this issue.
The Neural Tangent Generalization Attack (NTGA) is considered as the first well-known clean-label generalization attack under the black-box settings, which provided an unprecedented step in data protection approaches. In this paper, 
we conduct a comprehensive analysis on the state-of-the-art of NTGA; to the best of our knowledge, this is the first thorough analysis regarding NTGA. 
First, we provide a classification of 
attacks against DNNs 
with their explanations and relations to NTGA.
Then, this paper presents a taxonomy of black-box attacks and demonstrate 
that the NTGA is the first clean-label generalization attack under the black-box setting. 
We further analyze the existing studies of NTGA and give a comprehensive comparisons of their findings by conducting our own experiments to verify these findings.
Moreover, our extensive experiments show that NTGA is vulnerable to adversarial training and image transformations, and applying linear separability to NTGA-generated images makes them 
more susceptible to such vulnerablities. We present the pros and cons of NTGA and suggest ways to improve NTGA robustness based on our analysis.
Our further experiments indicate that several recently proposed clean-label generalization attacks outperform NTGA on data protection. 
Finally, we unveil the necessity of further research with future research insights on NTGA.
\end{abstract}
\ccsdesc[500]{Computing methodologies~Machine learning}
\ccsdesc[500]{Security and privacy}
\keywords{
clean-label generalization attacks, data protection approaches, Deep Neural Networks, Neural Tangent Generalization Attack}

\maketitle

\section{Introduction}

\begin{table*}[hbt!]
\normalsize
\caption{An overview of the current status of NTGA. The table summarizes the key findings of this study. We discuss these findings 
    in detail in Section~\ref{Sec:CurrentStatus}.  }
\renewcommand{\arraystretch}{1}
\setlength{\arrayrulewidth}{0.1mm}
    \begin{tabular}{|lp{6cm}p{9cm}|}
        \hline
        \rowcolor{black} \multicolumn{3}{|c|}{\textbf{\textcolor{white}{ Current Status of the NTGA}}}\\ 
        \hline\hline
        \rowcolor{lightgray}
        Section & Title & Key findings based on our analysis  \\ 
        \hline
        Section III A & Performance of the NTGA compared to other attacks & Some attacks such as error-minimizing, DeepConfuse attack, One Pixel Shortcut, and Synthetic outperforms the NTGA. However, the NTGA is able to challenge the error-maximizing attack and autoregressive attack.~\cite{DBLP:journals/corr/abs-2111-00898,DBLP:conf/iclr/FuHLST22,DBLP:conf/icml/YuanW21, DBLP:journals/cviu/ChenXZYXH25,DBLP:conf/aaai/LiuXC024}. 
        \\
        \hline
        Section III B  & Challenges against the NTGA & Adversarial training challenges the NTGA as well as other clean-label generalization
        attacks. Moreover, NTGA slightly degrade test robustness of adversarial training ~\cite{DBLP:conf/iclr/FuHLST22,DBLP:journals/corr/abs-2201-13329,sadasivan2023fun}.\\
         & 
        & NTGA is vulnerable to image transformation techniques such as matrix transformations~\cite{DBLP:journals/corr/abs-2111-13244,DBLP:journals/corr/abs-2301-13838,DBLP:journals/corr/abs-2303-08500,dataaugNTGA}.\\
        \hline
        Section III C & Linear separability of the NTGA  & NTGA-perturbed images and several other  following other clean-label generalization attacks are linearly separable~\cite{DBLP:journals/corr/abs-2111-00898,sadasivan2023fun}.\\
         \hline
         Section III D & Other NTK-based attacks against DNNs & Attacks based on NTKs~\cite{tsilivis2022the,DBLP:conf/icml/YuanW21} show strong transferability among models. \\ 
         \hline
    \end{tabular}
    \label{tb:currect_status}
\end{table*}

Because of their remarkable performance, Deep Neural Networks (DNNs) are used in many different applications of various areas~\cite{Goodfellow-et-al-2016,DBLP:conf/ccnc/LiLX21,DBLP:journals/arc/PoznyakCP19}. In order to achieve the best results out of DNNs, they need to be trained with a large set of data. On the other hand, the Internet 
is comprised of an enormous amount of data that belong to different owners. As a result of the desperate need for large datasets, DNNs practitioners tend to collect data from the Internet in an illegitimate manner. Hence, data owners who are unwilling to share their information are demanding methods to protect their data from unauthorized exploitation.


Fortunately, substantial research efforts have been carried out to avoid the unauthorized use of data through data protection approaches. In 2018, Yuan \& Wu \cite{DBLP:conf/icml/YuanW21} introduced such a data protection approach, the Neural Tangent Generalization Attack (NTGA), which is based on the Neural Tangent Kernels (NTKs) 
\cite{DBLP:conf/nips/JacotHG18}. NTGA is crafted by adding imperceptible perturbations to the training dataset. Then, the model trained on perturbed dataset will perform poorly on a validation or test dataset. Therefore, those data are worthless for DNN practitioners. Since the perturbations are imperceptible to the human eye, data can be used for other regular purposes as usual. 

NTGA represents a significant advancement in data protection approaches for several reasons: (1) it addresses the transferability of data protection approaches across models; (2) it leverages Gaussian Processes through the NTK approximation of wide neural networks to generate data poisoning attacks; and (3) it demonstrates remarkable performance in protecting data compared to existing approaches at the time.  Its major importance lies in being the first-ever clean-label generalization attack under \textit{the black-box setting}. Why is the black-box nature crucial? Because we cannot predict what model an unauthorized user might employ, the perturbations must be effective against any model. 
Seven years after the introduction of the NTGA, we revisit its role in the domain of data protection approaches. Our objective is to explicitly address the following question: \textit{What has been discovered about the NTGA since its proposal, and what future research directions does it suggest?} Even though there are substantial surveys on adversarial attacks and data poisoning attacks, which may lead the way to data protection approaches, none of them explicitly discuss or review the NTGA  
~\cite{DBLP:conf/dsc/FanYLQX22,ahmed2021threats,DBLP:journals/dcan/0011C0ML22,10.1145/3551636,DBLP:journals/compsec/LongGXZ22,DBLP:journals/corr/abs-2206-12227,DBLP:journals/corr/abs-2503-23536}.

Though NTGA is a data protection approach, it can also be viewed as an attack against DNNs. In this paper, we present a taxonomy of NTGA-related attacks against DNNs while providing fundamental knowledge on each type of attack. The taxonomy clearly positions NTGA within the broader landscape of attacks targeting DNNs. Specifically, NTGA is a generalization attack under the category of data poisoning, aiming to degrade model performance by corrupting the training data. Although generalization attacks have been studied for years, they face challenges when targeting diverse DNN architectures. NTGA addresses this issue by incorporating NTK theory. Due to its architecture-agnostic nature, NTGA can be classified as a black-box data poisoning attack, which is a rarely explored area. To support this fact, we present a separate taxonomy focused on black-box attacks against DNNs. 

In the literature, we found several studies that reveal essential features of NTGA from different perspectives which were not discussed in Yuan \& Wu~\cite{DBLP:conf/icml/YuanW21}. Section~\ref{Sec:CurrentStatus} is organized based on them. We provide a summary of this section in Table~\ref{tb:currect_status}. First, we explore the performance of NTGA compared to other clean-label generalization attacks (data protection approaches).  When introducing  NTGA, Yuan \& Wu~\cite{DBLP:conf/icml/YuanW21} only compared 
their attacks with DeepConfuse~\cite{DBLP:conf/nips/FengCZ19} and Return Favour Attack (RFA)~\cite{DBLP:journals/make/Chan-Hon-Tong19}. The standard practice when proposing a new attack is to compare the effectiveness of the proposed method with existing attacks. Hence, it is essential to identify the strong and weak attacks so that the researchers can easily decide which attacks to consider in their experiments. Our paper provides enough facts including experimental demonstrations for the researchers to make that decision on NTGA. Next, we analyze the current status of the NTGA against its  challenges, adversarial training, and data augmentation. Other than analyzing the existing studies, we provide our own experimental evidences to explain our insights and conclusions mentioned in Table~\ref{tb:currect_status}. After, we explore the linear separability of NTGA and other data protection approaches and how does it impact their robustness. We provide experimental results to show that an authorized user can determine the whether data are perturbed based on their linear separability. Furthermore, we investigate other attacks~\cite{DBLP:conf/nips/JacotHG18} that use Neural Tangent Kernels (NTKs), which serve as the building blocks of NTGA and identify their similarities with the NTGA. 

After evaluating the current status of NTGA, we conclude our findings by explicitly stating the pros and cons of the attack. These conclusions provide insights to facilitate researchers in making decisions regarding the use of NTGA. Then, we propose new insights and perspectives on NTGA that can be used for further research. New research directions include analyzing the transferability of NTGA with other attacks, applying NTGA to real-world data, investigating the performance of NTGA on distributed machine learning algorithms and unsupervised machine learning algorithms, and extending NTGA on other data types.

\begin{figure}[htb!]
\centering
    \includegraphics[scale=0.42]{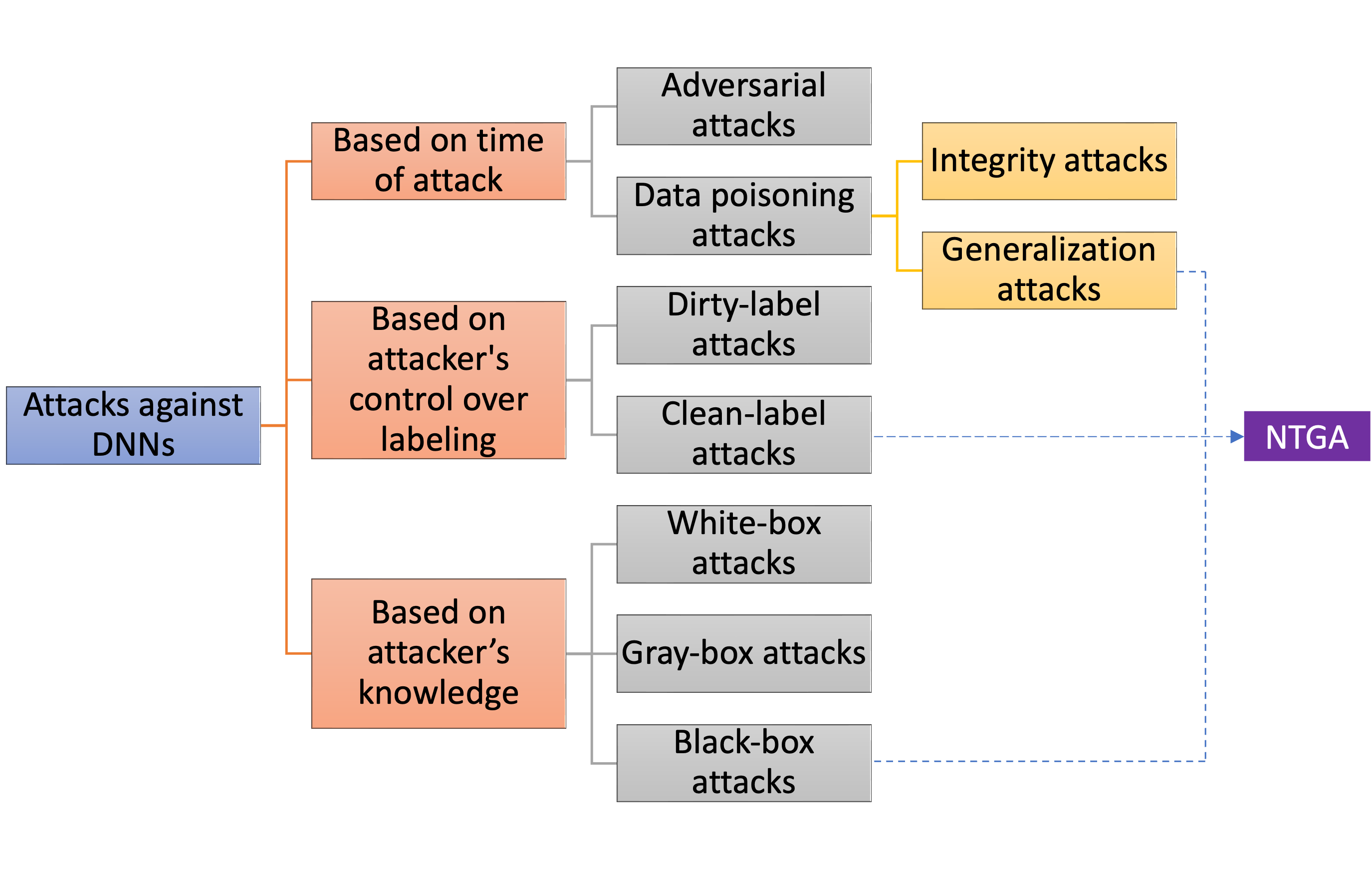}
    \caption{Attacks against DNNs can be categorized using several criteria. We consider three major criteria that are also relevant to NTGA. This classification illustrates the fact that NTGA belong to clean-label black-box generalization attack. }
    \label{fig:attacks_chart}
\end{figure}


The contributions of this paper can be summarized as follows:
\begin{itemize}
 \item{We specify where NTGA lies in the large domain of attacks against DNNs. In addition, we 
 delineate NTGA's role within the broader domain of attacks targeting DNNs. Moreover, we present a taxonomy of black-box attacks on DNNs to highlight the significance of NTGA among existing methods.}
 \item{We conduct a comprehensive analysis of the current status of NTGA and identify gaps not covered in Yuan \& Wu \cite{DBLP:conf/icml/YuanW21}. Our analysis reveals that several recently proposed data protection approaches outperform NTGA. Through experiments, we reveal NTGA’s susceptibility to adversarial training and data augmentation, and explore strategies for addressing these limitations. We also reveal that the linear separability of images perturbed by NTGA and related attacks increases their susceptibility to additional threats.}
 
 \item{Based on the challenges and key features of NTGA identified through our study, we propose several potential directions for future research on NTGA.}
 
\end{itemize}


\section{Background}
\label{Sec:Background}

We first discuss a classification of attacks against DNNs. Then, we focus our discussion on clean-label generalization attacks in this paper.  Besides NTGA, we also consider a few other clean-label generalization attacks. Furthermore, we study black-box attacks against DNNs because 
NTGA plays a vital role in the context of black-box attacks. Moreover, we expand our discussion to adversarial training, one of the most powerful defenses against prevailing attacks.

\subsection{Attacks against Deep Neural Networks (DNNs)}

Attacks against DNNs can be classified in different ways. Fig.~\ref{fig:attacks_chart} shows a particular classification of these attacks focusing on data protection approaches, where one of the main criteria for classifying DNN attacks is according to their time of occurrence. Based on the criterion, these attacks are classified into two categories as \textit{data poisoning attacks} and \textit{adversarial attacks}. 

\textit{Data poisoning attacks}: They occur during model training. Data poisoning attacks manipulate training data such that the model performs poorly on clean test data~\cite{DBLP:conf/iclr/00020000023}. These attacks corrupt DNNs and make them unable to be used for generalization purposes. Various data poisoning attacks have been introduced so far, such as the error-minimizing attack, the error-maximizing attack, NTGA, and so on \cite{DBLP:conf/nips/FowlGCGCG21}\cite{DBLP:conf/nips/FengCZ19}\cite{DBLP:conf/iclr/HuangME0021}\cite{DBLP:journals/corr/abs-2111-10130}\cite{DBLP:journals/corr/abs-2205-12141}. Data poisoning attacks also refer as \textit{causative attacks} \cite{DBLP:journals/corr/abs-2009-03728}\cite{DBLP:conf/icml/BiggioNL12}. We can further classify data poisoning attacks into two categories as generalization attacks and integrity attacks. Generalization attacks are also known as \textit{availability attacks} \cite{DBLP:conf/wacv/ZhaoL22}, and their main goal is to degrade overall model accuracy, including validation and test accuracy. On the other hand, integrity attacks cause the model to misclassify on specific images in a clean test set and degrade the test accuracy of the model. \textit{Poison Frogs} \cite{DBLP:conf/nips/ShafahiHNSSDG18} is one of the major integrity attacks \cite{DBLP:conf/nips/HuangGFTG20}\cite{DBLP:conf/icml/ZhuHLTSG19}. Unlike generalization attacks, integrity attacks do not mitigate validation accuracy.

\textit{Adversarial attacks}: While data poisoning attacks harm the model at training time, adversarial attacks occur at test time. Attackers add imperceptible perturbation to the test data so that pre-trained models misclassify them with high confidentiality during test time \cite{DBLP:journals/corr/SzegedyZSBEGF13}\cite{DBLP:journals/corr/GoodfellowSS14}. Adversarial attacks are also known as \textit{evasive attacks} \cite{DBLP:journals/corr/abs-2009-03728}. Popular adversarial attacks include Projected Gradient Descent (PGD) attack \cite{DBLP:journals/corr/MadryMSTV17}, Fast Gradient Sign Method (FGSM) attack~\cite{DBLP:journals/corr/HuangPGDA17}, DeepFool attack \cite{DBLP:conf/cvpr/Moosavi-Dezfooli16} and Carlini and Wagner’s attack (C\&W) \cite{DBLP:conf/sp/Carlini017}.

By definition, an adversarial attack is a mapping $\alpha: R^n \rightarrow R^n$ such that adversarial example $\alpha(x)=x'$ is misclassified as to a class other than its original class $y$ by the model $f$. The difference between $x'$ and $x$ is trivial, i.e., $\lVert x - x'\rVert_{p} \le \epsilon$ for some small value $\epsilon$ \cite{lin2021ml}. For a better understanding of adversarial attacks, as an example, we discuss the FGSM \cite{DBLP:journals/corr/HuangPGDA17} attack, where adversarial example $x'$ is crafted by solving the following optimization problem:
\begin{equation*}
    \label{eq:adv_attack}
\max_{x'}\mathcal{L}(f(x',y))
\end{equation*}
subject to
\begin{equation}
    \|x-x'\|_{\infty}\leq \epsilon,
\end{equation}
where $x'$ is generated by maximizing the loss function of DNN model $f$. At the same time, the attacker is trying to minimize the dissimilarity between the adversarial example and the original data. Hence, a trained model will misclassify $x'$  even though it looks like $x$, where a human can correctly classify it as class $y$.

Furthermore, we can categorize attacks against DNNs, considering if the attacker manipulates the input's target labels. If the attacker changes the target labels when building the attack, we call them {\it dirty-label attacks}. On the other hand, if the attacker is not influencing the labels of the inputs, those attacks are called {\it clean-label attacks}.
According to the attacker's knowledge regarding the target model, attacks can also be classified into three categories. 
(1) In \textit{black-box attacks}, the attacker has no information about the targeted model. (2) If the attacker has partial knowledge about the model, not exact information such as weights, those attacks are called \textit{gray-box attacks}. (3) \textit{White-box attacks} happen when the attacker has complete knowledge of the targeted model. However, the classification of attacks against DNNs is not restricted to these three criteria in Fig.~\ref{fig:attacks_chart} \cite{vorobeychik2018adversarial}\cite{DBLP:journals/dcan/0011C0ML22}. This paper only focuses on these three criteria relevant to NTGA. As illustrated in Fig.~\ref{fig:attacks_chart}, NTGA can be categorized as clean-label, black-box generalization attacks. 

Next, we thoroughly explore clean-label generalization attacks. Notations in Table.~\ref{tb:notations} in the Appendix
are helpful to refer the next sections.
Let a training set is denoted by $X^D$ and target labels of the training set is given by $Y^D$. A validation set is denoted by $X^V$ and labels of the validation set is given by $Y^V$. The set of perturbations given to $X^D$ is denoted by $\delta$.  Perturbations are bounded by the $L_p$ norm; i.e., $\|\delta\|_p \le \epsilon$. Note that $f$ denotes the target model and $\theta$ represent the set of model parameters. $\mathcal{L}$ represents the model's loss function.
\begin{definition}[Clean-label generalization attacks]

Clean-label generalization attack is crafted by solving the following bi-level optimization problem \cite{DBLP:conf/icml/YuanW21}\cite{DBLP:journals/corr/abs-2206-03693}\cite{DBLP:journals/make/Chan-Hon-Tong19}:
\begin{equation*}
    \label{eq:data_po1}
    \arg \max_{\|\delta\|_p \leq \epsilon} \left[ \mathcal{L}(f(X^V;\theta^*),Y^V)\right] \\
\end{equation*}
subject to
\begin{equation}
   \label{eq:data_po2}
    \theta^*=\arg\min_{\theta}\left[ \mathcal{L}(f(X^D +\delta;\theta),Y^D)\right]
\end{equation}
\end{definition} 

\begin{figure}[ht!]
    \centering
    \includegraphics[scale=0.45]{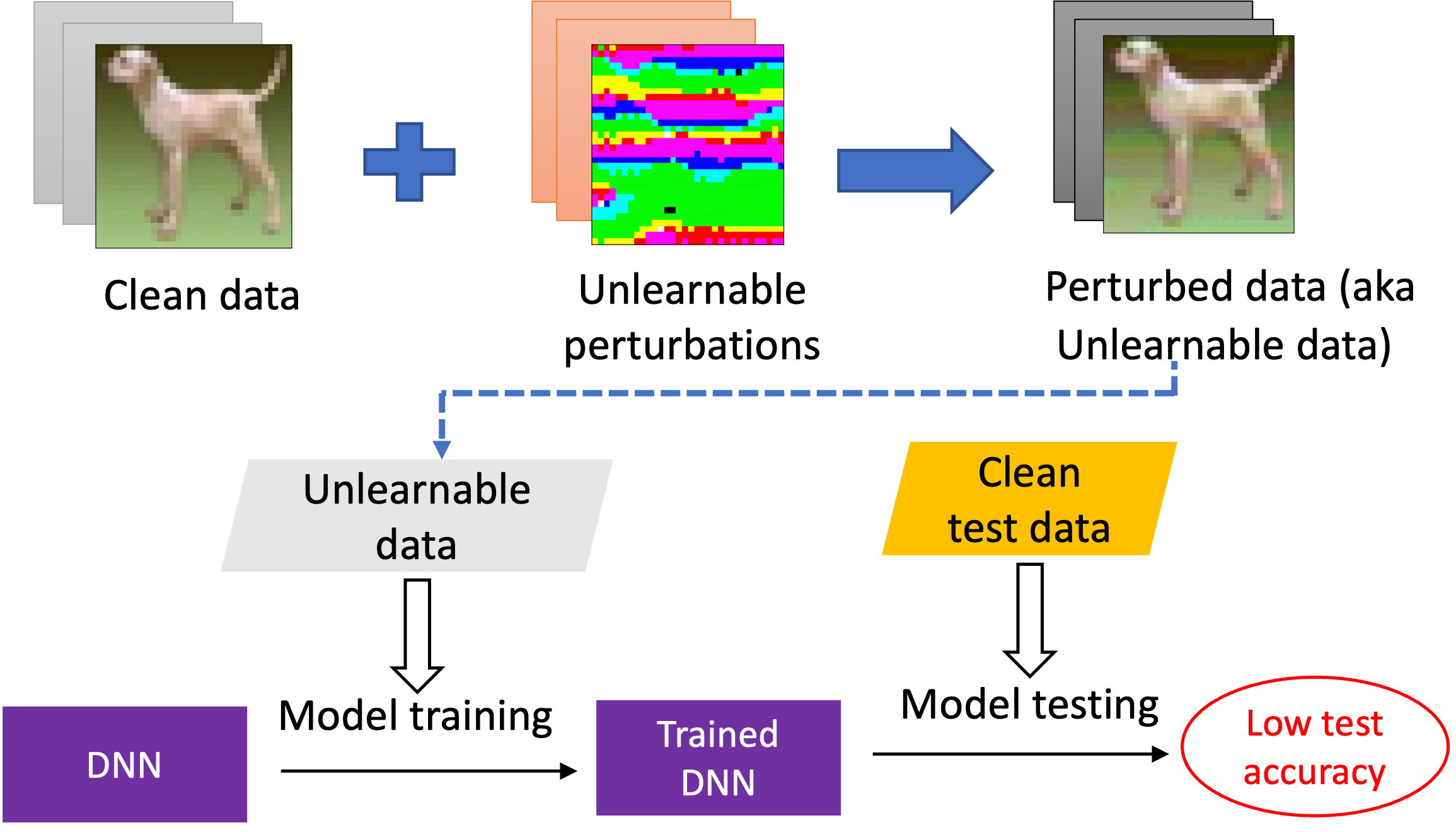}
    \caption{The function of perturbations generated by generalization attacks including NTGA.}
    \label{fig:NTGA_perturbations}
    \vspace{-0.5cm}
\end{figure}

As we explained before, generalization attacks perform poorly not only in 
a test dataset but also in 
a validation set. Fig.~\ref{fig:NTGA_perturbations} illustrates the function of perturbations generated by generalization attacks. \textcolor{black}{There are two groups: a \textit{data owner} and an \textit{unauthorized user}. As shown at the top of Fig.~\ref{fig:NTGA_perturbations}, the data owner adds imperceptible perturbations to the original data. These perturbations are generated through clean-label generalization attacks. The perturbed data are also referred to as \textit{unlearnable data}.
An unauthorized user uses the perturbed data to train a DNN model. The model may achieve high training accuracy; however, when it is used to make predictions on unseen data without perturbations (i.e., test/validation data), it produces incorrect predictions. This is because the DNN model cannot effectively learn from the training data due to the unlearnable perturbations. Therefore, the data become useless for an unauthorized user when training models.}
The data owner crafts the perturbations by minimizing the training loss and 
maximizing the validation loss. Under black-box setting $f$ in Equation~\ref{eq:data_po1} is unknown. $f$ can be any unknown target model decided by unauthorized person. The NTGA is crafted by representing $f$ using mean of Gaussian processes along with the NTK. Next, we explain NTGA. 

\subsection{Neural Tangent Generalization Attacks (NTGA)}

It is a known fact that infinite width neural networks and Gaussian processes 
are equivalent at initialization \cite{DBLP:journals/corr/abs-1711-00165}. In a fully connected neural network, when the limit of widths of hidden layers goes to infinity, the network function at initialization converges to a Gaussian distribution related to a kernel. Moreover, it is identified that during the training of neural network, the evolution of a neural network function can be described by the kernel: NTK \cite{DBLP:conf/nips/JacotHG18}. 

\begin{definition} The Neural Tangent Kernel (NTK) between two inputs $x_i$ and $ x_j$ coming from a neural network with an infinite width limit and random initialization converges to the following kernel function \cite{DBLP:conf/nips/JacotHG18}:
\begin{equation}
    K(x_i,x_j)=\mathbb{E}_{\theta} \nabla_{\theta}f(x_i;\theta)^T\nabla_{\theta}f(x_j;\theta).
\end{equation}
\end{definition}

As seen, a kernel function is related to the above covariance function. The NTK between two inputs refers to the similarity of the two input data. Hence, an approximation to the range of neural networks can be established combining the two concepts \cite{DBLP:conf/nips/JacotHG18}. As a result, we can rewrite the neural network function $f$ in Equation~\ref{eq:data_po1} using Gaussian processes, along with the NTK. Then, the maximization problem in Equation~\ref{eq:data_po1} can be modified as~\cite{DBLP:conf/icml/YuanW21}:
\begin{equation}
    \label{eq:NTGA}
    \arg\max_{\|\delta\|_p \leq \epsilon} \left[ \mathcal{L}(\Bar{f}(X^V;\hat{K}^{V,D},\hat{K}^{D,D},Y^D,t),Y^V)\right],
\end{equation}
where $\Bar{f}$ denotes the mean of Gaussian processes, $\hat{K}^{D,D}$ is the kernel matrix of training dataset, and $\hat{K}^{V,D}$ is the  kernel matrix between validation dataset and training dataset. The hyperparameter $t$ controls the time of attack happens.  The NTGA is crafted by solving the above optimization problem using projected gradient ascent.

Throughout our study, we involve several other clean-label generalization attacks to understand the current status of NTGA among its competitors. We briefly discuss those attacks in the following section for a better understanding.

\subsection{Other Clean-label Generalization Attacks}
\label{Sec:app3}
\label{Sec:otherattacks}
\subsubsection{Error-maximizing attack (Adversarial poisoning)}
The perturbed images generated by adversarial attacks are known as adversarial examples \cite{DBLP:journals/corr/GoodfellowSS14}. Fowl et al. \cite{DBLP:conf/nips/FowlGCGCG21} revealed that adversarial examples could be more effective as data poisoning attacks. They used a different approach from the usual data poisoning attack formulated in Definition~\ref{eq:data_po1}. Simply, they applied adversarial attacks mechanism on a fixed pre-trained model to generate poisons. This approach can be formulated as follows:
\begin{equation}
    \label{eq:emax}
    \arg \max_{\|\delta\|_p \leq \epsilon} \left[ \mathcal{L}(f(X^D+\delta; \theta^*),Y^D) \right].
\end{equation}
In contrast to NTGA, error-maximizing attacks do not follow a bi-level optimization process. $\theta^*$ in Equation~\ref{eq:emax} represents the model parameters trained over clean data, and it is fixed during the generation of perturbations. We can see that this optimization problem is the same as the Equation~\ref{eq:adv_attack}, which is used to craft adversarial attack. Fowl et al. \cite{DBLP:conf/nips/FowlGCGCG21} generated perturbations by solving this optimization problem using Projected Gradient Descent (PGD). 

\subsubsection{Error-minimizing attack}
Error-minimizing attack makes data unlearnable for DNNs by adding a type of imperceptible noise based on the opposite direction of the optimization problem of the error-maximizing attack. This noise is generated in a way such that the error of one or more training examples is made closer to zero. Therefore, it tricks the model by indicating that there is nothing to learn from these examples. Huang et al. \cite{DBLP:conf/iclr/HuangME0021} presented the following bi-level optimization problem to generate the attack.
\begin{equation}
\label{eq:emin}
    \arg\min_\theta\left[ \min_{\|\delta\|_p \leq \epsilon} \mathcal{L}(f(X^D+\delta;\theta),Y^D) \right].
\end{equation}
Note that both optimization problems in Equation~\ref{eq:emin} have the same objective of minimizing the loss. The inner optimization problem finds perturbations $\delta$ that minimize the training loss. The outer optimization problem finds the parameters that minimize the same loss. In this case, they optimize $\delta$ over the training set after every M steps of optimizing $\theta$. Huang et al.~\cite{DBLP:conf/iclr/HuangME0021} solved the inner minimization problem using PGD.


\subsubsection{DeepConfuse attack}

DeepConfuse is another data poisoning attack that we encounter throughout this paper. When solving optimization problem in~\ref{eq:data_po1},  Feng et al.~\cite{DBLP:conf/nips/FengCZ19} used a encoder-decoder neural network $g_{\xi}$ to generate the noise. Hence, we can rewrite the Eq.~\ref{eq:data_po1} as below.  

\begin{equation*}
    \arg \max_{\epsilon} \left[ \mathcal{L}(f(X^D;\theta^*),Y^D)\right] \\
\end{equation*}
subject to
\begin{equation}
   \label{eq:dc}
    \theta^*=\arg\min_{\theta}\left[ \mathcal{L}(f(X^D+g_{\xi}(X^D);\theta),Y^D)\right].
\end{equation}
Feng et al.~\cite{DBLP:conf/nips/FengCZ19} alternatively updated $f$ on the perturbed data using the gradient descent method and updated the weights of $g_{\xi}$ on the clean data using gradient ascent.

\subsection{Adversarial Training}
\label{app:2}
Adversarial training is a powerful remedy for most attacks, including data protection approaches such as NTGA. Adversarial training is training DNNs on adversarial examples so that the models will not be vulnerable to adversarial attacks at test time. Adversarial examples are crafted using adversarial attacks such as PGD attacks \cite{DBLP:journals/corr/MadryMSTV17}. 
Even though adversarial training is initially introduced to defend against adversarial attacks, it is shown that adversarial training can also be used against data poisoning attacks \cite{DBLP:conf/iclr/FuHLST22}\cite{geiping2022what}\cite{DBLP:journals/corr/abs-2111-10130}. Mathematically, adversarial training can be formulated using optimization problem:
\begin{equation}
    \label{eq:AT}
    \min_{\theta} \left[\max_{\|\delta\|_p \leq \rho_a} \mathcal{L}(f(X^D+\delta),Y^D;\theta)\right], 
\end{equation} where $\rho_a > 0$ denotes the adversarial perturbation radius \cite{DBLP:journals/corr/MadryMSTV17} which is the maximum perturbation allowed. Sometimes, adversarial perturbation radius is also called a defense budget \cite{DBLP:journals/corr/abs-2201-13329}. A larger perturbation radius will give stronger adversarial robustness, but it will result in more visible perturbations. The inner maximization problem in Equation~\ref{eq:AT} finds a perturbation for each input that gives maximum loss, and the outer optimization problem finds the model parameters that minimize the adversarial loss in the inner problem. We can see that adversarial training allows a certain perturbation limit for each input while training. Therefore, the model cannot be tricked by adversarial inputs and training time perturbations \cite{DBLP:conf/nips/HuangGFTG20}.


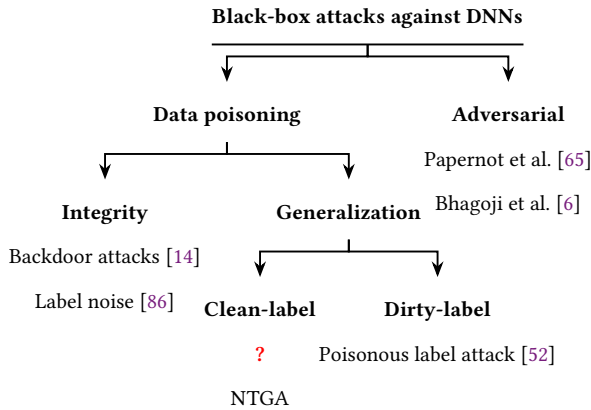
\begin{figure}[ht!]
\centering
\scalebox{0.9}{ 
\begin{forest}
  for tree={
    align=center,
    edge+={thick, -{Stealth[]}},
    l sep+=5pt,        
    s sep=-5pt,        
    fork sep=5pt,      
  },
  forked edges,
  if level=0{
    inner xsep=0pt,
    tikz={\draw [thick] (.children first) -- (.children last);}
  }{},
  [\textbf{Black-box attacks against DNNs} \\
    [\textbf{Data poisoning}
        [\textbf{Integrity} \\Backdoor attacks~\cite{DBLP:journals/corr/abs-1712-05526}\\Label noise~\cite{DBLP:conf/nips/RooyenMW15},]
        [\textbf{Generalization}
          [\textbf{Clean-label}\\
           \textbf{\textcolor{red}{?}}\\NTGA]
          [\textbf{Dirty-label}\\ Poisonous label attack~\cite{DBLP:journals/npl/LiuLL21}]
          ]
       ]   
    [\textbf{Adversarial}\\Papernot et al.~\cite{DBLP:conf/ccs/PapernotMGJCS17}\\ Bhagoji et al.~\cite{DBLP:journals/corr/abs-1712-09491}]
  ]
\end{forest}
}
\vspace{-0.6cm}
\caption{The taxonomy of attacks under the black-box setting is displayed. By exploring example attacks under each type of black-box attack, we identify that there is a lack of black-box attacks under clean-label generalization attacks. We explicitly show how NTGA are fulfilling that gap in the taxonomy.}
\label{fig:BBattacks}
\vspace{-0.3cm}
\end{figure}

\subsection{What make NTGA different from other data protection approaches?}

The main difference between NTGA and other clean-label generalization attacks mentioned in Section~\ref{Sec:otherattacks} is being a black-box attack. Applying Gaussian processes and NTKs to approximate the entire wide neural networks makes NTGA a successful black-box attack. Other data protection approaches described in~\ref{Sec:otherattacks} use pre-determined models when generating the attack, which may result in the lack of transferability of their perturbations. Yuan \& Wu \cite{DBLP:conf/icml/YuanW21} claimed that the NTGA 
is the first clean-label generalization attack under the black-box setting, to which we agree. Even though there are several black-box integrity attacks, such as back-door attacks \cite{DBLP:journals/corr/abs-1712-05526} and Label noise \cite{DBLP:conf/nips/RooyenMW15}, black-box generalization attacks are rarely explored. \textit{Poisonous label attack} \cite{DBLP:journals/npl/LiuLL21} is a black-box generalization attack, but unlike NTGA, it is a dirty-label attack. MetaPoison \cite{DBLP:conf/nips/HuangGFTG20} and Poison Frogs \cite{DBLP:conf/nips/ShafahiHNSSDG18} are clean-label integrity attacks but cannot be considered as black-box attacks. However, there are several black-box attacks in the context of adversarial attacks, which we will not thoroughly explore in this paper~\cite{DBLP:conf/ccs/PapernotMGJCS17}\cite{DBLP:journals/corr/abs-1712-09491}. Fig.~\ref{fig:BBattacks} clearly shows a visual depiction taxonomy of attacks against DNNs under black-box settings. Using taxonomy in Fig.~\ref{fig:BBattacks},  we verify that NTGA fulfill the lack of black-box attacks under clean-label generalization attacks.


Other data protection approaches, such as watermarking~\cite{sharma2024review} and membership inference attacks~\cite{DBLP:journals/corr/ShokriSS16}, have different objectives from NTGA. Watermarking is designed to embed identifiable information such as, ownership or authentication marks into images, whereas NTGA aims to corrupt the training process of a DNN model and reduce its generalization ability. Specifically, watermarking is not intended have a negative impact on a DNN model’s performance. In contrast, membership inference attacks aim to determine whether a particular image was included in a model’s training dataset, allowing data owners to verify if their data has been used for training a specific model. NTGA, on the other hand, takes a different approach: it allows others to use the protected data for training, but the generated noise corrupts the learning process, making the data ineffective for training purposes. As a result, users are discouraged from using such data.

\section{Current Status of NTGA}
\label{Sec:CurrentStatus}

In this section, we elaborate on the current status of NTGA from five aspects. None of these perspectives were discussed in Yuan \& Wu \cite{DBLP:conf/icml/YuanW21} when introducing NTGA. At the end of the section, we present a summary of the pros and cons of NTGA.

\subsection{Performance of NTGA compared to Other Attacks}
\begin{table}[ht!]
\renewcommand{\arraystretch}{1}
\setlength{\arrayrulewidth}{0.1mm}
\normalsize
    \centering
    \vspace{-0.2cm}
    \caption{Test accuracy of a pretrained ResNet50 trained on CIFAR-10 protected by each data protection approaches.}
    \begin{tabular}{|l|l|}
    \hline
        Attacks & Test Accuracy (\%) \\
    \hline
        NTGA~\cite{DBLP:conf/icml/YuanW21} & 55.32\\
        \hline
        DeepConfuse~\cite{DBLP:conf/nips/FengCZ19} & \textbf{40.21} \\
        Error-minimizing~\cite{DBLP:conf/iclr/HuangME0021} & \textbf{30.76} \\
        Error-maximizing~\cite{DBLP:conf/nips/FowlGCGCG21} & 76.56 \\
        Synthetic~\cite{DBLP:journals/corr/abs-2111-00898} & \textbf{47.30} \\
        Robust Error-Minimizing~\cite{DBLP:conf/iclr/FuHLST22} & \textbf{52.88} \\
        One-Pixel shortcut~\cite{DBLP:journals/corr/abs-2205-12141} & \textbf{32.21} \\
        Autoregressive~\cite{DBLP:journals/corr/abs-2206-03693} & 57.95 \\
        \hline
    \end{tabular}
    \label{tab:performance}
    \vspace{-0.4cm}
\end{table}
Yuan \& Wu~\cite{DBLP:conf/icml/YuanW21} compared NTGA with the DeepConfuse~\cite{DBLP:conf/nips/FengCZ19} attack and the Return Favour Attack (RFA)~\cite{DBLP:journals/make/Chan-Hon-Tong19}, which were the baselines of their study. According to their results, the NTGA outperforms both the DeepConfuse attack and the RFA in a black-box setting but not in a gray-box setting. Later, several studies proposed new attacks and compared the attacks directly with NTGA in performance. Those studies revealed further information about the performance of NTGA.

Yu et al.~\cite{DBLP:journals/corr/abs-2111-00898} studied the performance of several data protection approaches, including NTGA. Their results shows that the error-minimizing attack, the error-maximizing attack, synthetic perturbations and the DeepConfuse attack perform better than the NTGA based on the test accuracy. Furthermore, by comparing the test accuracies reported in Fu et al.~\cite{DBLP:conf/iclr/FuHLST22}, it was shown that the error-minimizing attack outperforms NTGA, while the error-maximizing attack does not. The Robust Error-Minimizing (REM) approach introduced by Fu et al.~\cite{DBLP:conf/iclr/FuHLST22} appears to offer a slight improvement over NTGA in terms of data protection.
\textcolor{black}{NTGA demonstrated performance comparable to the recently introduced Enhanced Unlearnable Examples (EUN)~\cite{DBLP:journals/cviu/ChenXZYXH25}, which are generated using convolutional operations and single pixel-level modifications. However, unlike EUN, NTGA exhibits performance degradation when models are trained with adversarial training. A similar vulnerability is observed in NTGA against Stable Error-Minimizing noise (SEM)~\cite{DBLP:conf/aaai/LiuXC024}, another recently proposed data protection approach. SEM modifies the REM approach by training defensive noise against random perturbations instead of adversarial perturbations. Overall, while NTGA is competitive with current data protection approaches, its vulnerability to adversarial training remains a significant limitation.
} 

\textcolor{black}{However, some recently proposed approaches, such as autoregressive perturbations~\cite{DBLP:journals/corr/abs-2206-03693} and the One-Pixel shortcut attack~\cite{DBLP:journals/corr/abs-2205-12141}, have not been directly compared with NTGA. Moreover, when reviewing multiple studies, it is difficult to draw definitive conclusions about their performance. For instance, Yu et al.~\cite{DBLP:journals/corr/abs-2111-00898} reported that the error-maximizing attack outperforms NTGA, whereas Fu et al.~\cite{DBLP:conf/iclr/FuHLST22} found that NTGA performs better than the error-maximizing approach. To further investigate, we conducted our own comparison by training a ResNet50 model on the CIFAR-10 dataset protected by various data protection methods. Refer to Appendix~\ref{app:2} for further details on the experimental settings. The results are presented in Table~\ref{tab:performance}. Bold values indicate test accuracies lower than those achieved by NTGA, implying that those methods perform better. Based on these results, NTGA outperforms both the error-maximizing and autoregressive approaches. However, these conclusions are highly dependent on the dataset and model architecture, and a comprehensive analysis is necessary to identify their performance under different circumstances.}

Another challenge of NTGA, as pointed out by Sandoval-Segura et al.~\cite{DBLP:journals/corr/abs-2206-03693}, is that NTGA take a long time to generate the perturbations or noises, making it difficult to apply NTGA to large datasets. This is caused by involving a surrogate model when crafting perturbations. Sadasivan et al.~\cite{DBLP:journals/corr/abs-2303-04278} showed that the NTGA took 5.2 hours to generate noises while error-maximizing attacks took only about 0.5 hours under the same experimental settings.


\vspace{-0.4cm}
\subsection{Challenges against NTGA}

\subsubsection{\textbf{Adversarial Training}}


In this section, we discuss the performance of NTGA against adversarial training~\cite{DBLP:journals/corr/MadryMSTV17}. 
Fu et al. \cite{DBLP:conf/iclr/FuHLST22} shows that major data protection approaches are challenged by adversarial training. Especially after realizing that error-minimizing attacks can be defeated by adversarial training, Fu et al. \cite{DBLP:conf/iclr/FuHLST22} proposed an extended version of error-minimizing attacks called \textit{the Robust Error-Minimizing (REM)} attack that is robust to adversarial training. 
Their study demonstrates that adversarial training cannot defeat REM attacks while other baseline attacks fail to do so. Their baseline attacks include the NTGA, the error-minimizing attack, and the error-maximizing attack.

The behavior of NTGA under adversarial training can be further understood by Tao et al.~\cite{DBLP:journals/corr/abs-2201-13329} which introduces the stability attack, a clean-label attack that challenges the test-time robustness of adversarial training. According to the results in Tao et al.~\cite{DBLP:journals/corr/abs-2201-13329}, we can observe that the an adversarially trained model on NTGA gives almost the same test accuracy as an adversarially trained model  clean data after. These results implies that adversarial training defeats the protection given by the NTGA. Hence, it confirms the results of Fu et al.~\cite{DBLP:conf/iclr/FuHLST22}: the NTGA cannot protect data against adversarial training. However, data poisoning attacks such as One-Pixel Shortcut~\cite{DBLP:journals/corr/abs-2205-12141}, ADVIN attacks~\cite{DBLP:journals/corr/abs-2111-10130}, REM~\cite{DBLP:conf/iclr/FuHLST22}, Fang et al.~\cite{fang2024re} have been able to challenge adversarial training. Furthermore, Sadasivan et al.~\cite{sadasivan2023fun} introduced a data protection approach called, Filter-based UNlearnable (\textit{FUN}), which provides evidence that the protection, given by the NTGA, the error-minimizing attacks, and the error-maximizing attacks on the CIFAR-10 dataset, can be eliminated using adversarial training with a perturbation radius of 4/255. Moreover, Sadasivan et al.~\cite{DBLP:journals/corr/abs-2303-04278} introduced a convolution-based data protection approach (CUDA), which is a faster and more effective approach against adversarial training. In that study, they confirmed that the NTGA, as well as error-minimizing, and error-maximizing attacks can be defeated by adversarial training with a perturbation of 4/255.

As we mentioned in Section~\ref{Sec:Background}, adversarial training is initially designed against adversarial attacks. That means adversarial training is supposed to defend the models against attacks that occur at test time. Tao et al.~\cite{DBLP:journals/corr/abs-2201-13329} investigated the effectiveness of adversarial training conducted on protected data under test-time attacks. In other words, they observed the test time robustness of adversarially trained models on protected data. First, Tao et al.~\cite{DBLP:journals/corr/abs-2201-13329} conducted adversarial training (PGD-AT~\cite{DBLP:journals/corr/MadryMSTV17}) on data protected by different approaches such as the DeepConfuse, the NTGA and the error-maximizing attack. Then, they create test datasets with different noises to evaluate the test robustness of adversarially trained models. Test-time adversarial attacks such as PGD~\cite{DBLP:journals/corr/MadryMSTV17}, FGSM~\cite{DBLP:journals/corr/HuangPGDA17}, and C\&W~\cite{DBLP:conf/sp/Carlini017} are used to generate noises. 
We noticed that adversarial training conducted on clean data gives the test accuracy of around 50\% against adversarial attacks. However, adversarial training, conducted on data protected by the NTGA, results in slightly less test accuracies than the clean data. The reduction in the robust accuracies implies that the NTGA degrades the test-robustness of adversarial training slightly. However, DeepConfuse and the error-maximizing attacks are mitigating test-robustness better than the NTGA. The stability attack proposed by Tao et al.~\cite{DBLP:journals/corr/abs-2201-13329} outperforms all the clean-label generalization attacks considered  in terms of degrading the test-robustness of adversarial training.

\begin{figure}[ht!]
    \centering
    \includegraphics[width=\linewidth]{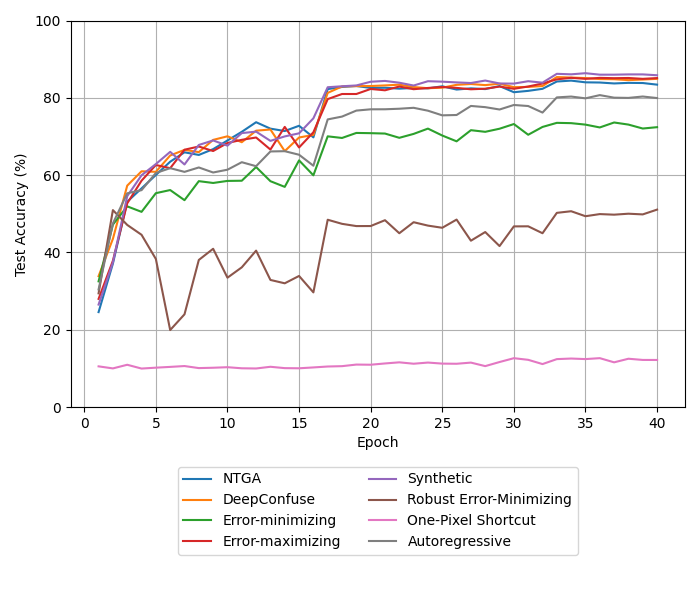}
    \vspace{-1cm}
    \caption{Test accuracy curves when adversarial training is conducted on protected CIFAR-10 datasets.}
    \label{fig:adv_trn}
    \vspace{-0.5cm}
\end{figure}

In Yuan \& Wu~\cite{DBLP:conf/icml/YuanW21}, they have not tested NTGA against adversarial training.  \textcolor{black}{The information discussed in the preceding three paragraphs leads to the conclusion that NTGA may be vulnerable to adversarial training. To verify this suspicion, we have conducted adversarial training on CIFAR-10 datasets protected by eight different data protection approaches. The base model used was VGG19, adversarially trained using PGD attacks  with a perturbation radius of 4/255 and a step size of 0.8/255. The resulting test accuracies are presented in Table~\ref{tab:AT}. A clean dataset typically yields a test accuracy of around 85\%. As shown in Table~\ref{tab:AT}, NTGA, DeepConfuse, Error-Minimizing, Error-Maximizing, and Autoregressive approaches have reached similar accuracy levels, indicating their vulnerability to adversarial training. Moreover, our experiments have shown that One-Pixel Shortcut and REM are more robust to adversarial training. Fig.~\ref{fig:adv_trn} clearly illustrates the robustness of One-Pixel Shortcut and REM to adversarial training  using the test accuracy curves obtained during adversarial training. This highlights a potential direction for improving NTGA, as adversarial training remains a significant challenge. For instance, the REM approach generates noise using an adversarially trained model. Similarly, NTGA could be enhanced by incorporating an adversarially trained surrogate model instead of relying on standard training. Additionally, shortcut learning techniques from the One-Pixel Shortcut method, where only one pixel was modified per image class, could be integrated to further strengthen NTGA against adversarial attacks.} 

\begin{table}[ht!]
\renewcommand{\arraystretch}{1}
\setlength{\arrayrulewidth}{0.1mm}
\normalsize
    \centering
    \caption{Test accuracy of a VGGG19 model trained on CIFAR-10 dataset protected by each data protection approaches.}
    \begin{tabular}{|l|l|}
    \hline 
        Attacks & Test Accuracy (\%) \\
    \hline
        NTGA~\cite{DBLP:conf/icml/YuanW21} & 83.41\\
        DeepConfuse~\cite{DBLP:conf/nips/FengCZ19} & 84.91 \\
        Error-minimizing~\cite{DBLP:conf/iclr/HuangME0021}  & 72.41 \\
        Error-maximizing~\cite{DBLP:conf/nips/FowlGCGCG21} & 85.11 \\
        Synthetic~\cite{DBLP:journals/corr/abs-2111-00898} & 85.84 \\
        Robust Error-Minimizing~\cite{DBLP:conf/iclr/FuHLST22} & 51.09 \\
        One-Pixel Shortcut~\cite{DBLP:journals/corr/abs-2205-12141}  & 12.20  \\
        Autoregressive~\cite{DBLP:journals/corr/abs-2206-03693}  & 79.94 \\
        \hline
    \end{tabular}
    \label{tab:AT}
    \vspace{-0.5cm}
\end{table}

\subsubsection{\textbf{Data Augmentation}}
It is a known fact that strong data augmentation can defeat data poisoning attacks~\cite{DBLP:journals/corr/abs-2301-13838,DBLP:journals/corr/abs-2111-13244}. Huang et al.~\cite{DBLP:conf/iclr/HuangME0021} claimed that the error-minimizing attacks is robust to strong data augmentation techniques, such as CutMix~\cite{DBLP:conf/iccv/YunHCOYC19}, Cutout~\cite{DBLP:journals/corr/abs-1708-04552} and mixup~\cite{DBLP:conf/iclr/ZhangCDL18}. However, Liu et al.~\cite{DBLP:journals/corr/abs-2111-13244} showed that the effect of error-minimizing attacks can be mitigated by simple grayscale pre-filtering. Moreover, Borgnia et al.~\cite{DBLP:conf/icassp/BorgniaCFGGGGG21} diminished the effect of backdoor attacks~\cite{DBLP:journals/corr/abs-1712-05526} using mixup~\cite{DBLP:conf/iclr/ZhangCDL18} and CutMix~\cite{DBLP:conf/iccv/YunHCOYC19} data augmentation techniques. Hence, data poisoning attacks have unpredictable reactions to data augmentation which needs to be explored further. Moreover, certain data augmentation techniques used for image compression such as JPEG compression, SHIELD~\cite{DBLP:conf/kdd/DasSCHLCKC18} provide defense against adversarial attacks~\cite{DBLP:journals/corr/abs-2012-01701}\cite{DBLP:conf/ica3pp/Zeng0MQ20}. 

Liu et al.~\cite{DBLP:journals/corr/abs-2301-13838} reveals that many data protection approaches including NTGA are also vulnerable to JPEG compression and grayscale transformations.
A recent study~\cite{dataaugNTGA}
showed that some clean-label generalization attacks, including the NTGA, can be defeated by nonlinear transformations. Once a training dataset is increased based on simple yet effective nonlinear transformations techniques, the NTGA can no longer protect the data. Unlike Fu et al.~\cite{DBLP:conf/iclr/FuHLST22}, they did not incorporate any adversarial training techniques. In Hapuarachchi et al.~\cite{dataaugNTGA}, we used data augmentation techniques such as erode~\cite{article}, dilate~\cite{article}, color channel manipulation, and Gaussian blur~\cite{gedraite2011investigation}.  
However, as shown in their study, this weakness is common to all clean-label generalization attacks described in Section~\ref{Sec:otherattacks}, not only NTGA. \textcolor{black}{On the other hand, Sandoval-Segura et al.~\cite{DBLP:conf/nips/SeguraSGGG23} performed linear transformations to attack data protected by the availability-based methods, including NTGA. They captured the added linearly separable perturbations and attempted to remove them from the data. Their approach improved test accuracy on the ResNet18 model trained on the NTGA-perturbed CIFAR-10 dataset from 40.78\% to 82.21\%, showing that even linear transformations can effectively compromise NTGA.}

Dolatabadi et al.~\cite{DBLP:journals/corr/abs-2303-08500} proposed a new mechanism to defeat data protection approaches called \textit{dAta aVailAbiliTy Attacks defuseR} (\textit{AVATAR}) that was inspired by data augmentation. They used Gaussian noise over the protected training data and conducted reverse diffusion using a pre-trained diffusion model to remove the perturbations in the data. In the study, they showed that data protection approaches such as NTGA, the error-minimizing attack, and the error-maximizing attack are vulnerable to \textit{AVATAR}. The \textit{AVATAR} did not incorporate the data augmentation techniques used by Hapuarachchi et al.~\cite{dataaugNTGA}. However, Dolatabadi et al.~\cite{DBLP:journals/corr/abs-2303-08500} showed that their approach performs better than the data augmentation techniques, such as Cutout, CutMix, and mixup. Moreover, Yu et al.~\cite{DBLP:conf/icml/YuWXYLTK24} proposed a method to remove perturbations introduced by data protection approaches, including NTGA, using autoencoders. They claim that their approach outperforms AVATAR in attacking NTGA. 

Countermeasures for data augmentation-based attacks have also been explored in the literature, e.g.,~\cite{DBLP:journals/corr/abs-2301-13838,DBLP:journals/corr/abs-2501-08862}. Gong et al.~\cite{DBLP:journals/corr/abs-2501-08862} proposed a method to generate perturbations that are robust against data augmentation. The perturbation generation process involves a carefully designed surrogate model and an augmentation selection strategy. Additionally, Liu et al.~\cite{DBLP:journals/corr/abs-2301-13838} applied adaptive poisoning to error-minimizing and Synthetic attacks to counter the effects of data augmentation techniques such as grayscale. However, they did not consider NTGA, which we identify as a gap worth exploring in future to develop more robust attacks against data augmentation.

\begin{figure}[ht!]
    \centering
    \vspace{-0.3cm}
    \includegraphics[width=\linewidth]{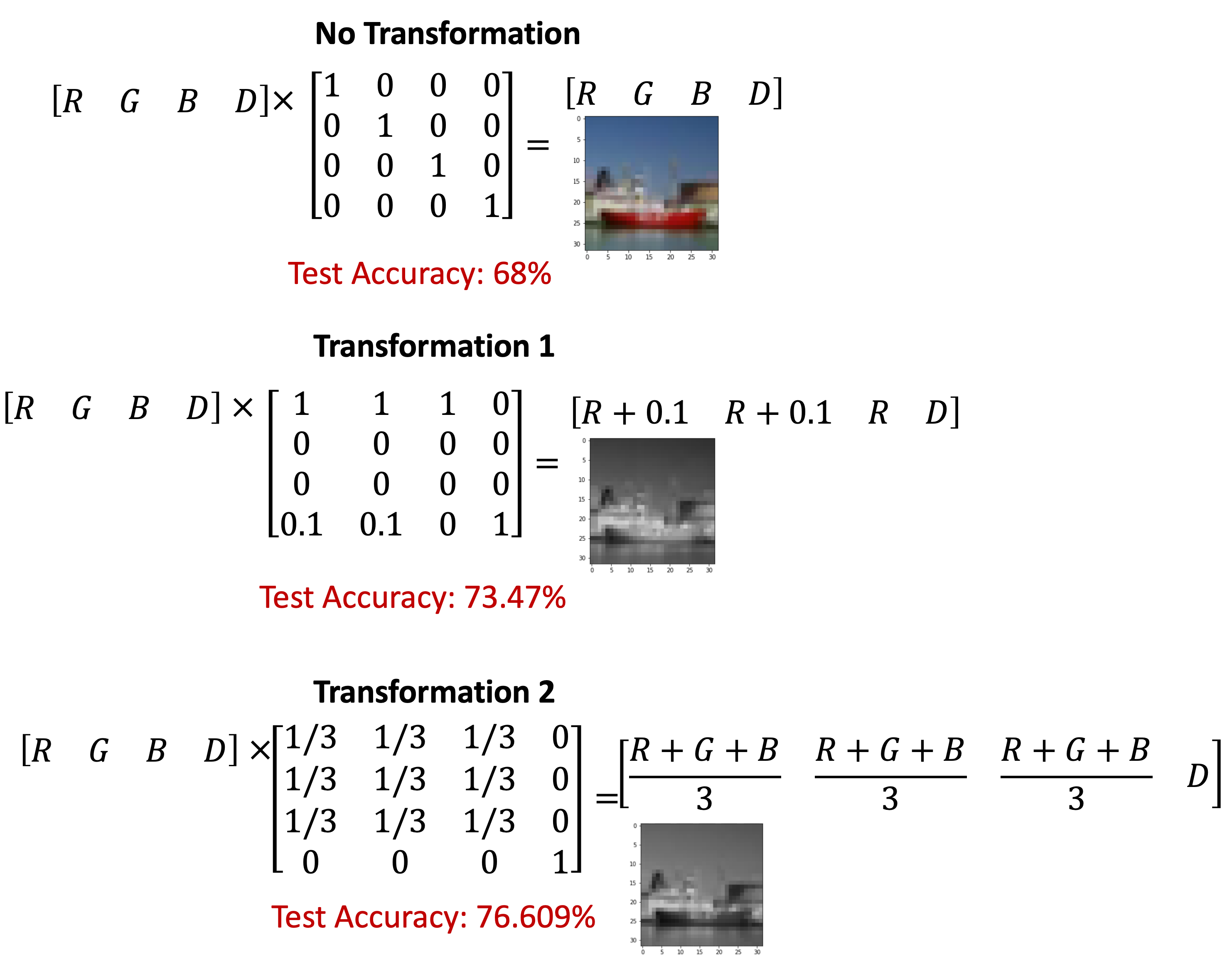}
    \vspace{-0.7cm}
    \caption{Matrix transformations on NTGA-perturbed data. The variables \textit{R}, \textit{G}, and \textit{B} represent the red, green, and blue channel values of each pixel, respectively, while \textit{D} is a dummy variable introduced to enhance the transformation.}
    \label{fig:transformation}
    \vspace{-0.3cm}
\end{figure}

In the above analysis, we observed that grayscale and a similar transformation known as color channel manipulation significantly affected the performance of the data protection approaches including NTGA~\cite{DBLP:journals/corr/abs-2111-13244,DBLP:journals/corr/abs-2301-13838,dataaugNTGA}. Similarly, identifying a specific transformation that strongly impacts perturbations could be valuable for enhancing the robustness of the NTGA. Although transformation techniques have been widely investigated, prior work has not focused on pinpointing a single definitive transformation with this effect. A promising starting point is the transformation proposed in~\cite{transform}, which closely resembles the grayscale transformation. This method implements the transformation by applying matrix multiplication to the image data. Our objective is to identify a transformation matrix that most effectively improves the accuracy of the test. Figure~\ref{fig:transformation} illustrates how such a transformation can be applied to NTGA-perturbed data. The variables \textit{R}, \textit{G}, and \textit{B} represent the red, green, and blue channel values of each pixel, respectively, while \textit{D} is a dummy variable introduced to enhance the transformation. As shown in Figure~\ref{fig:transformation}, we modified the transformation matrix with the goal of discovering an optimal configuration that can neutralize NTGA perturbations and test accuracy. 
Identifying a transformation that consistently disrupts NTGA perturbations would represent a significant advancement toward more robust data protection methods.

\subsection{Linearly Separability of NTGA} 

Yu et al.~\cite{DBLP:journals/corr/abs-2111-00898} unveiled that perturbations generated by clean-label generalization attacks, including the NTGA, are linearly separable. They considered the DeepConfuse attack, the error-minimizing attack, the error-maximizing attack, and the NTGA to prove their argument. First, Yu et al.~\cite{DBLP:journals/corr/abs-2111-00898} created two-dimensional t-SNE plots of perturbations and clean data. The t-SNE plot is a statistical method to display high-dimensional data points in a two or three-dimensional plot~\cite{JMLR:v9:vandermaaten08a}. Surprisingly, they observed that perturbations with a same class are clustered together, and clean data did not exhibit such a pattern. Sadasivan et al.~\cite{sadasivan2023fun} also observed a similar pattern in the t-SNE plot of REM~\cite{DBLP:conf/iclr/FuHLST22} perturbations. Furthermore, Yu et al.~\cite{DBLP:journals/corr/abs-2111-00898} confirmed the linear separability by fitting linear models on perturbations. Additionally, they showed that simple logistic regression models between target classes and perturbations can be fitted with more than 90\% training accuracy. They further proved their argument using a one-layer (linear) neural network and a two-layer neural network to classify perturbations. The linear model achieved more than 90\% training accuracy, while the two-layer network gave nearly 100\% training accuracy for all attacks considered. Hence, these results confirmed that all clean-label generalization attacks considered are linearly separable.


\begin{table}[ht!]
\renewcommand{\arraystretch}{1}
\setlength{\arrayrulewidth}{0.1mm}
\normalsize
    \centering
    \caption{Training accuracy of the logistic regression model fitted on CIFAR-10 datasets.}
    \begin{tabular}{|l|l|}
    \hline 
        Attacks & Training Accuracy (\%) \\
    \hline
        Clean & 48.04\\
        NTGA~\cite{DBLP:conf/icml/YuanW21}  & 99.79\\
        DeepConfuse~\cite{DBLP:conf/nips/FengCZ19} & 98.99 \\
        Error-minimizing~\cite{DBLP:conf/iclr/HuangME0021}  & 99.98 \\
        Error-maximizing~\cite{DBLP:conf/nips/FowlGCGCG21} & 76.12 \\
        Synthetic~\cite{DBLP:journals/corr/abs-2111-00898} & 100 \\
        Robust Error-Minimizing~\cite{DBLP:conf/iclr/FuHLST22} & 77.82 \\
        One-Pixel Shortcut~\cite{DBLP:journals/corr/abs-2205-12141}  & 100  \\
        Autoregressive~\cite{DBLP:journals/corr/abs-2206-03693} & 48.75 \\
        \hline
    \end{tabular}
    \label{tab:logistic}
    \vspace{-0.3cm}
\end{table}

Furthermore, as demonstrated by Yu et al.~\cite{DBLP:journals/corr/abs-2111-00898}, linear separability is a sufficient condition for availability attacks to succeed. They proved their argument by using the simple synthetic data as perturbations. First, they generated synthetic perturbations and further processed them to defend image augmentations and showed that synthetic perturbations 
outperformed most of the attacks. Hence, Yu et al.~\cite{DBLP:journals/corr/abs-2111-00898} have confirmed that the linear separability is a sufficient condition for generalization attacks. \textcolor{black}{However, Sandoval-Segura et al.~\cite{DBLP:conf/nips/SeguraSGGG23} challenged the notion that data protected by data protection approaches are effective due to linearly separable perturbations. They presented autoregressive perturbations~\cite{DBLP:journals/corr/abs-2206-03693} as a counterexample, which do not exhibit linear separability, unlike other attacks such as NTGA that exhibits the linear separability property. To investigate this property, they trained a linear logistic regression model on the perturbations to assess their linearly separability. Furthermore, Zhu et al.~\cite{DBLP:conf/aaai/ZhuYG24} confirmed that NTGA is indeed linearly separable and developed a detection algorithm capable of identifying data protected by availability attacks based on this property.}

\textcolor{black}{In the aforementioned studies, researchers analyzed the linear separability of perturbations, not of the perturbed images. For instance, Yu et al.\cite{DBLP:journals/corr/abs-2111-00898} created t-SNE plots on perturbations using a linear (one-layer) neural network, while Sandoval-Segura et al.\cite{DBLP:conf/nips/SeguraSGGG23} applied logistic regression models on perturbations. However, none of these studies evaluated the linear separability of the perturbed images themselves. In a realistic setting, an unauthorized user may only have access to perturbed data. Such a user could attempt to exploit the linear separability property of the perturbed data, but they cannot directly extract the perturbations since they lack access to the clean data. To study this scenario, we conducted experiments to evaluate whether perturbed images exhibit linear separability. Specifically, we fit a logistic regression model with the L-BFGS solver on the CIFAR-10 dataset, using the images and their corresponding class labels. The training accuracies of the fitted models are reported in Table~\ref{tab:logistic}. Our results show that images perturbed with NTGA, DeepConfuse, error-minimizing attack, and One-Pixel Shortcut exhibit strong linear relationships with the class labels, while Error-Maximizing and REM perturbations show moderate relationships. In contrast, confirming the findings of Sandoval-Segura et al.~\cite{DBLP:conf/nips/SeguraSGGG23}, autoregressive perturbations do not display such linear separability; their training accuracy is nearly identical to that of clean data. This indicates that an unauthorized user cannot manipulate or detect autoregressive perturbations through linear separability. Overall, this analysis suggests that NTGA can be improved by reducing the linear separability of its perturbations. }

\subsection{Other Attacks against DNNs using NTKs}

Tsilivis \& Kempe,~\cite{tsilivis2022the} proposed an adversarial attack using the NTK, called \textit{NTK perturbations}. Besides NTGA, this is the only prevailing attack involving the NTK. Tsilivis and Kempe~\cite{tsilivis2022the} showed that NTK perturbations perform almost the same as the PGD attack~\cite{DBLP:journals/corr/MadryMSTV17}, one of the most powerful adversarial attacks. Another advantage of NTK perturbations over NTGA is, providing an analytical expression for the perturbations found using the NTK as follows~\cite{tsilivis2022the}.
\begin{equation}
    \label{eq:NTK_Per}
    \eta_i = -\epsilon y_i \cdot sgn(A_i^TH(X,X)^{-1}Y)
\end{equation} where $\eta_i$ denotes the perturbation bounded by $\epsilon$. 
$H$ represents the NTK matrix. $X$ and $Y$ indicate the training set and target labels, respectively. $A_i$ denotes the matrix including derivatives of the NTK between $x_i$ and $x_j$ for all $j=1\dots n$. 

It is interesting to notice that both NTGA and NTK perturbations show strong transferability among models. It means that both NTGA and NTK perturbations derived from a specific model can fool other types of models successfully. We can suggest that involving NTKs may have contributed to the transferability of their attacks. Thus, it is worth exploring the possibility of incorporating NTKs in order to create successful black-box attacks. 

To conclude this section, we provide Table~\ref{tb:ProsandCons}, which summarizes the pros and cons of the NTGA.
\begin{table}[htb!]
\caption{A summary of the pros and cons of NTGA.}
\renewcommand{\arraystretch}{1}
\setlength{\arrayrulewidth}{0.1mm}
    \begin{tabular}{|p{4.0cm}|p{4.0cm}|}
        \hline
        Pros & Cons  \\ 
        \hline
        Establish new 
         state-of-the-art for clean-label generalization attacks under black-box settings. & Several other data protection approaches perform better than the NTGA.  \\
        \hline
        Remarkable transferability among a wide range of DNNs. & Cannot protect data against adversarial training. \\
        \hline
        Capable of protecting data better than some other prevailing data protection approaches. & Take long time to generate the attack since a surrogate model is involved. \\
        \hline
        Introduced under both black-box and gray-box settings. & Can be defeated by simple image transformation techniques. \\
        \hline
    \end{tabular}
    \label{tb:ProsandCons}
    \vspace{-0.5cm}
\end{table}

\section{Future Research Insights}
\label{sec:future_work}

\subsection{Transferability of NTGA Compared to Other Clean-label Generalization Attacks}

We have noticed that the primary significance of NTGA, among other clean-label generalization attacks, is being a black-box attack. The property of transferability guides the way to a successful black-box attack~\cite{DBLP:journals/corr/LiuCLS16}. The transferability of perturbations means that perturbations derived from a specific model can mislead models with different architectures. Even though several other data protection approaches outperform the NTGA, more studies need to be conducted to compare the transferability of NTGA with other generalization attacks. Since black-box attacks are primarily introduced in adversarial attack settings, testing for transferability can be found mainly in the context of adversarial attacks~\cite{DBLP:journals/corr/PapernotMG16}. 

A common approach for testing the transferability of data protection methods is generating perturbed images using a certain model architecture, training a model with a different architecture using those images, and investigating the trained model's generalization ability. Furthermore, there are advanced practices for evaluating transferability. One approach is utilizing machine learning services hosted by Google and Amazon. Huang et al.~\cite{DBLP:conf/nips/HuangGFTG20}, which proposed a clean-label data poisoning attack called MetaPoison, examined the transferability using the Google Cloud AutoML API~\cite{GoogleML}. Given a dataset, Google Cloud AutoML API~\cite{GoogleML} allows us to train models in a black-box setting, hiding the training architecture from the user. Papernot \& McDaniel~\cite{DBLP:conf/ccs/PapernotMGJCS17} used both Google and Amazon machine learning services to experiment transferability~\cite{AWS}. Moreover, Google Cloud Vision API~\cite{googleAPI} is a tool that can be used to evaluate adversarial attacks in a black-box setting~\cite{DBLP:journals/corr/abs-1911-07140}.

We suggest testing the transferability of other clean-label generalization attacks, such as the error-maximizing~\cite{DBLP:conf/iclr/HuangME0021}, and the error-minimizing~\cite{DBLP:conf/nips/FowlGCGCG21} attacks, to emphasize the performance of NTGA compared to others. \textcolor{black}{As a method for improving transferability, we can also consider the approach proposed by Chen et al.~\cite{DBLP:conf/iclr/ChenYC0QWH23}. They generated an attack using model checkpoints that exhibit diverse behaviors, effectively simulating DNNs with different architectures, rather than relying on a single model.}

\subsection{Applications of Clean-label Generalization Attacks to Real-world Datasets}

Throughout our study, we observed that the generalization attacks protect data effectively based on the experiments conducted using the benchmark datasets, such as the CIFAR-10~\cite{krizhevsky2009learning}, the ImageNet~\cite{DBLP:conf/cvpr/DengDSLL009} and the MNIST~\cite{DBLP:journals/corr/abs-1102-0183}. However, the possibility of applying these attacks on real-life data is yet to be investigated. Moreover, it is vital to verify that the effectiveness of these attacks prevails on the real-world datasets. Given that the real-world data are more complicated than the standard datasets, exploring the compatibility of these attacks under realistic settings is essential. Huang et al.~\cite{DBLP:conf/iclr/HuangME0021} provided such a demonstration of their attack on the WebFace dataset~\cite{DBLP:journals/corr/YiLLL14a}, a widely used face recognition dataset. Furthermore, Wang et al.~\cite{DBLP:journals/corr/abs-2111-10130} demonstrated a real-world application of the ADVIN attack on the same dataset. Shan et al.~\cite{DBLP:conf/uss/ShanWZLZZ20}, proposed a data protection approach against the facial recognition DNN models based on the WebFace~\cite{DBLP:journals/corr/YiLLL14a} and the VGGFace2~\cite{DBLP:conf/fgr/CaoSXPZ18} datasets. 

Moreover, it would be beneficial to provide guidelines on applying these attacks in real-life data so that any data owner can easily utilize them to protect their data. Ma et al.~\cite{DBLP:journals/tvcg/MaXLM20} extended his study to provide a visual analytics framework to show model vulnerabilities to data poisoning attacks. In the literature, we can observe that some adversarial attacks are evaluated in real-world data~\cite{DBLP:conf/iclr/KurakinGB17a}\cite{DBLP:conf/ccs/SharifBBR16}. For instance, Kurakin et al.~\cite{DBLP:conf/iclr/KurakinGB17a} examined the vulnerability of adversarial images obtained from mobile phone cameras. However, there is a lack of data poisoning attacks that are being evaluated under realistic scenarios. The concept used for adversarial attacks must be restructured for data poisoning attacks since they are executed during training, not in testing. \textcolor{black}{Additionally, researchers can leverage APBench~\cite{qin2023apbench}, which offers accessible implementations of availability attacks and establishes a unified benchmark to support future research. By using APBench, inconsistencies in coding environments across different datasets and experimental setups can be mitigated, enabling more standardized and reproducible evaluations of various attack methods.}

\subsection{NTGA on Distributed Machine Learning}

Conventional machine learning algorithms are not efficient enough to handle the current rapid growth in data. Training larger datasets requires an enormous the number of parameters that is impossible to control with available computing power. However, given the fact that increasing data reduces the learning error, it is necessary to find a way to deal with large-scale data. Distributed machine learning is an approach that can handle extensive datasets analysis efficiently. Distributed machine learning algorithms are implemented on multiple nodes, which allows handing larger input datasets~\cite{galakatos2018distributed}\cite{DBLP:journals/csur/VerbraekenWKKVR20}. Since these approaches are prevalent, it is interesting to explore how NTGA perform under distributed machine learning. We can focus on the problem: \textit{Can NTGA protect data against distributed learning algorithms?} for future research.
Several studies have been conducted on data poisoning attacks under distributed machine learning in the past~\cite{tian2021defending}\cite{tomsett2019model}\cite{DBLP:journals/corr/abs-1808-04866}. They mainly involve federated learning~\cite{tian2021defending}, a popular variant of distributed machine learning.

\subsection{NTGA on Unsupervised Learning Models }

It is important to notice that NTGA are crafted and tested focusing only supervised learning algorithms. However, we cannot predict the unauthorized user's choice of the target model. Since unsupervised learning is also popular as supervised learning in deep learning applications, there is a valid chance that data would be trained on unsupervised learning models. Being a black-box attack, it is interesting to investigate how NTGA perform on unsupervised learning methods. Moreover, He et al.~\cite{DBLP:journals/corr/abs-2202-11202} motivates us to explore this research direction further. They showed that the error-minimizing attack~\cite{DBLP:conf/iclr/HuangME0021} and the error-maximizing attack~\cite{DBLP:conf/nips/FowlGCGCG21} could not protect data against unsupervised contrastive learning~\cite{DBLP:conf/cvpr/ChopraHL05} models, which is one of the most powerful unsupervised learning methods. \textcolor{black}{Moreover, Wang et al.~\cite{DBLP:conf/nips/WangZG24} showed that NTGA were ineffective in mitigating the performance of the SimCLR model~\cite{chen2020simple}, a contrastive learning framework.}  In the literature, it is vital to notice that few studies have been done on data protection methods against unsupervised learning methods~\cite{DBLP:journals/corr/abs-2202-11202}.

\subsection{NTGA on Ensemble Learning Algorithms}
Another aspect we can explore is the behavior of NTGA under the ensemble learning. Ensemble learning~\cite{DBLP:journals/fcsc/DongYCSM20} methods utilize multiple learning algorithms with weak predictive results and combine results using voting mechanisms to achieve better performances than that obtained from any traditional algorithm alone. Traditional ensemble learning mechanisms include stacking, boosting, and bagging~\cite{DBLP:journals/fcsc/DongYCSM20}. In recent years, ensemble learning mechanisms have dominated machine learning applications, for example, the medical data analysis and the fraud detection~\cite{DBLP:journals/cogcom/WenHLLJX17}\cite{DBLP:journals/widm/SagiR18}\cite{DBLP:journals/cmpb/XiaoWLZ18}. Because of the impressive performance in ensemble learning methods, it is essential to explore whether NTGA can protect data against such advanced machine learning methods. Hapuarachchi et al.~\cite{hapuarachchi2025advancing} conducted a comprehensive analysis of leveraging ensemble learning with data generated through generalization attacks, including NTGA. They demonstrated that, with slight modifications, ensemble learning can be successfully applied on data protected by generalization attacks, revealing its vulnerabilities. 
We suggest investigating ways to improve NTGA’s robustness against such advanced ensemble learning algorithms. For instance, instead of relying on a single surrogate model when generating perturbations, NTGA could incorporate ensemble machine learning algorithms. 

\subsection{NTGA for Other Data Types}

\textcolor{black}{NTGA 
has primarily been introduced for image data So far. However, data protection approaches are now being extended to other domains such as text and audio~\cite{DBLP:conf/mm/LiuJXLC24,li2023make,wang2024pointapa}. Given the growing prominence and popularity of Natural Language Processing (NLP) and text-based applications, exploring how these methods can be adapted to protect textual data is extremely important. Some studies have already begun investigating this direction~\cite{DBLP:conf/mm/LiuJXLC24,li2023make}. For instance, Liu et al.~\cite{DBLP:conf/mm/LiuJXLC24} applied error-minimizing attacks to both images and their associated captions, aiming to protect multimodal content. The potential novelty of applying NTGA in this context lies in their black-box attack capabilities. Similarly, Li et al.~\cite{li2023make} adopted error-minimizing attacks for text data. Additionally, Gokul et al.~\cite{gokul2024poscuda} explored unlearnable examples in the audio domain using the CUDA~\cite{DBLP:journals/corr/abs-2303-04278}  framework.}




\section{Conclusion}
\label{Sec:Conclusion}
NTGA is undoubtedly a vital discovery in the area of clean-label generalization attacks. After seven years of proposing the attacks, this paper analyzes the current status of the NTGA among other clean-label generalization attacks. We have noticed that adversarial training and simple image transformations mitigate the protection given by NTGA, which is a typical case for other generalization attacks as well. Moreover, we have observed that major clean-label generalization attacks outperform NTGA in several cases. However, given that it is the only prevailing black-box clean-label generalization attack, it is worth to study for further improvement. Hence, we provide several insights for future research on NTGA based on the areas that have yet to explore. 


\bibliographystyle{ACM-Reference-Format}
\bibliography{references}

\section{Appendix}
\subsection{Notations}
\label{app:1}
\begin{table}[H]
\caption{Notation }
\renewcommand{\arraystretch}{1}
\setlength{\arrayrulewidth}{0.1mm}
\normalsize
\centering
    \begin{tabular}{|l|l|}
        \hline
        \rowcolor{lightgray} \multicolumn{2}{|c|}{{Notation}}\\ 
        \hline \hline
        $X^D$ & Training set  \\
        \hline
        $Y^D$ & Target labels of training set \\
        \hline
         $X^V$ & Validation set  \\
        \hline
         $Y^V$ & Target labels of validation set  \\
        \hline
        $\delta$ & Set of perturbations\\
        \hline
        $\epsilon$ & Maximum perturbation allowed\\
        \hline
        $f$ & Target model\\
        \hline
        $\theta$ & Model parameters\\
        \hline
        $\mathcal{L}$ & Model's lost function \\
        \hline
    \end{tabular}
    \label{tb:notations}
\end{table}

\subsection{Experimental Settings}
\label{app:2}

The NTGA dataset was collected from the data released by the authors in their GitHub repository~\cite{DBLP:conf/icml/YuanW21}. Error-minimizing, error-maximizing, DeepConfuse, and synthetic data were collected from the authors of Yu et al.~\cite{DBLP:journals/corr/abs-2111-00898}. Robust Error-Minimizing, One-Pixel Shortcut, and Autoregressive datasets were created using code from their GitHub repositories with default settings. According to Yu et al.~\cite{DBLP:journals/corr/abs-2111-00898}, DeepConfuse used an 8-layer U-Net as the surrogate model, while error-minimizing and error-maximizing used ResNet-18 models. NTGA used a CNN surrogate model. Robust Error-Minimizing employed a ResNet-18 surrogate model. Synthetic and One-Pixel Shortcut do not involve surrogate models. Perturbation budgets followed the default or the best conditions mentioned in the paper~\cite{DBLP:journals/corr/abs-2111-00898}. 

In our experiments, we used the SGD optimizer with 0.9 momentum.  The batch size was set to 100, and training was conducted for 30 epochs with a learning rate of 0.001. The loss function is cross-entropy loss. We did not use data augmentation techniques during training, as they can impact the protection given by data protection approaches.

\end{document}